\documentclass{article}
\usepackage{spconf,epsfig}
\usepackage[table,xcdraw]{xcolor}
\usepackage{graphicx}
\usepackage[cmex10]{amsmath}
\usepackage{algorithm}
\usepackage{algpseudocode}
\usepackage{hyperref}
\usepackage{parskip}
\usepackage{tikz}
\usepackage[colorinlistoftodos]{todonotes}
\usepackage{amsfonts}
\usepackage{amssymb}

\pdfoutput=1

\title{Semi-Supervised Learning With Graphs: Covariance Based Superpixels for Hyperspectral Image Classification}
\makeatletter
\def\@name{\emph{Philip~Sellars$^{1}$, Angelica I. Aviles-Rivero$^{2}$, Nicolas Papadakis$^{3}$,} \\{\emph{David Coomes$^{4}$, Anita Faul$^{5}$ and Carola-Bibiane Sch{\"o}nlieb}$^{1}$}}
\address{$^1$ DAMTP and $^2$DPMMS, Faculty of Mathematics, University of Cambridge, UK. \\
$^3$  IMB, Universit{\'e} Bordeaux France, $^4$ Department of Plant Science, University of Cambridge, UK.\\
$^5$ Laboratory for Scientific Computing, University of Cambridge, UK.
}

\newcommand\copyrighttext{%
	\footnotesize \textcopyright 2019 IEEE. Personal use of this material is 
	permitted.
	Permission from IEEE must be obtained for all other uses, in any current or 
	future
	media, including reprinting/republishing this material for advertising or 
	promotional
	purposes, creating new collective works, for resale or redistribution to 
	servers or
	lists, or reuse of any copyrighted component of this work in other works.}

\newcommand\copyrightnoticenew{%
	\begin{tikzpicture}[remember picture,overlay]
	\node[anchor=south,yshift=10pt] at (current page.south) 
	{\fbox{\parbox{\dimexpr\textwidth-\fboxsep-\fboxrule\relax}{\copyrighttext}}};
	\end{tikzpicture}%
}

\begin{document}
%\ninept
%
\maketitle
\copyrightnoticenew
\begin{abstract}
\text{ }\\
In this paper, we present a graph-based semi-supervised framework for hyperspectral image classification. We first introduce a novel superpixel algorithm based on the spectral covariance matrix representation of pixels to provide a better representation of our data. We then construct a superpixel graph, based on carefully considered feature vectors, before performing classification. We demonstrate, through a set of experimental results using two benchmarking datasets, that our approach outperforms three state-of-the-art classification frameworks, especially when an extremely small amount of labelled data is used.  
\end{abstract}

\section{Introduction}
Hyperspectral image (HSI) classification is an active area of research and poses unique challenges. The high dimensional nature of the data allows for a detailed description of an image, and class labels can be assigned on a pixel-by-pixel basis. The majority of classification frameworks for HSIs are supervised learning (SL) frameworks; during the training phase information is only gained from the initial labelled data. These include kernel methods \cite{SCMK}, deep-learning \cite{DRNN2017} and sparse representation methods \cite{SparseDictionary}. However, the problem with SL methods is that they rely upon the existence of a large and accurately labelled training set. In application such as HSI classification, label collection is time consuming and expensive. 

Another set of algorithmic approaches are based on unsupervised learning frameworks. However, the intrinsic nature of the problem makes the classification task a strongly ill-posed problem, and therefore, specific assumptions are needed to mitigate the lack of correspondence between the produced clusters and the known classes.

In practice, the size of the labelled set is often very small compared to the amount of unlabelled data. In such applications, there are large advantages to using semi-supervised learning (SSL) approaches \cite{SSTheory}. SSL methods use information present in the labelled and unlabelled data during the training process, which can lead to much higher classification accuracy. These approaches can be divided into three groups: generative, low-density separation and graph-based methods.

This paper follows the graph perspective. This is motivated by the advantages of using graphs: i)  a natural representation for HSI data, ii) a way to gain scalability and therefore computational tractability and  iii) a structure with mathematical desirable properties (e.g. sparseness).  However, a fundamental problem in graph based learning is \textit{how to construct the graph in order to produce an accurate classification?} This is of great interest as the performance of the classifier heavily depends on the feature selection and the structure design, and this question is addressed in this paper.

\textbf{Contributions.} We present a novel graph base framework for HSI 
classification, which we call hyperspectral superpixel graph classification 
(HSGC). Our approach achieves state-of-the-art classification results using a 
semi-supervised graph based approach alongside a carefully designed feature 
space. Our highlights are: 1) spectral covariance based superpixels for feature 
extraction which uses local covariance matrix representation to include spatial 
and spectral information, 2) a graphical representation where each node 
represents a superpixel and the edges represent the similarity being the 
superpixel feature vectors and 3) a unified SSL framework. We show that 
learning with minimal supervision is highly beneficial when one removes feature 
space redundancy whilst strengthening the synergy between feature selection and 
graph construction.
\vspace{-1mm}

\section{Learning With Minimal Supervision}
\label{sec:method}
This section is divided into three key parts. Firstly, we introduce our proposed algorithmic approach to segment HSIs. Secondly, we describe our feature extraction process. Finally, we construct the graphical representation and perform the classification task. 

\subsection{Spectral Covariance Based Superpixels}
Let $\textbf{I} = \{I_{b}\}, b= 1,..,\mathcal{B}$ be a HSI with dimensions $\mathcal{W} \times \mathcal{H} \times \mathcal{B}$ representing the width,  height and number of bands respectively and $I_{b}:\mathcal{W}\times \mathcal{H}\rightarrow D $ where $D$ is the image representation of a band. Our framework starts by performing dimensionality reduction, via PCA~\cite{PCA}, on $\textbf{I}$ for computational efficiency. We construct a dimensionally reduced HSI $\mathbf{\widehat{I}} = \{\widehat{I}_{a}\}, a= 1,..,\mathcal{A}$  where $\mathcal{A} \ll \mathcal{B}$.

We then aim to find a better representation of the HSI data to increase the performance of our classifier. This is an important step as classification accuracy is highly dependent on setting relevant local regions. Local regions are commonly set by using either a fixed size or dynamic window. However, this provides a limited representation. An alternative is to use superpixels, which has been explored  in~\cite{SCMK}. Unlike \cite{SCMK} which relied on an existing superpixel algorithm, we propose a new superpixel approach that is designed with HSIs in mind. 

Denoting an individual pixel as $p \in \mathbf{\widehat{I}}$, superpixels split the HSI into a family of disjoint sets, $ \mathbf{\widehat{I}} = \cup_{i=1}^{K} \mathcal{S}_i$ , $\mathcal{S}_i \cap \mathcal{S}_j = \emptyset $, where  $\mathcal{S}_i$ corresponds to an individual superpixel and $K$ is the number of superpixels, which is initially set by the user. Each superpixel $\mathcal{S}_i$ is made up of a set of $n_i$ connected pixels, $\mathcal{S}_i =  \{ p_{i,1} , ... , p_{i,n_i}  \}$. Our superpixel segmentations are produced via minimisation of the following objective,
%\begin{equation}
  $  Q({\{\mathcal{S}_1,..,\mathcal{S}_K \}}) = \sum_{i=1}^{K}\sum_{p \in \mathbf{\widehat{I}}} d((p,\mathbf{\widehat{I}(p)}),F(\mathcal{S}_i))$
%\end{equation}
where $d$ is a distance function and $F(\mathcal{S}_i)$ is the average of $\mathcal{S}_i$. 

We have built our superpixel algorithm on top of the commonly used SLIC algorithm~\cite{slic}. SLIC has drawbacks~\cite{2018arXiv180202796M} found in the fixed size localised search range. To improve upon this, we adopt the observation of~\cite{2018arXiv180202796M}, in which the search range is dynamically adjusted by the local content density in the image. This information is given by the function $g$ which maps each pixel to a positive real number. Therefore, in our superpixel algorithm, we used the following search range\begin{center}

$ d((p,\widehat{I}(p)),F(\mathcal{S}_i))\text{ if }|p-(F(\mathcal{S}_i))_1|\leq 2 \sqrt{\frac{\mathcal{W}\mathcal{H}}{K}} g(F(S^{}_i));$\\
$\text{Otherwise }d((p,\widehat{I}(p)),F(\mathcal{S}_i))=\infty\text; $
\end{center}
where $(F(\mathcal{S}_i))_1$ represents that spatial part of the feature function, and $|\cdot|$ is the euclidean distance on the image grid.

Furthermore, instead of using the Euclidean spectral distance found in ~\cite{slic,2018arXiv180202796M}, we instead use covariance matrix representation~\cite{CMR} and the Log-Euclidean distance (LED)~\cite{mbdm} which is better suited for HSIs. For each pixel ${p} \in \mathbf{\widehat{I}}$ we construct a covariance matrix $\textbf{C}_{{p}}$ describing the relationship between different hyperspectral bands, which extracts powerful spectral and spatial information. However, covariance matrices are symmetric positive definite matrices and they do not lie on a Euclidean space but instead on a Riemannian manifold. Therefore, the LED metric is used to construct the spectral distance between pixels,
\begin{equation}
    d_{spectral}(p_x,p_y) = || \text{logm} (\textbf{C}_{p_x}) - \text{logm} (\textbf{C}_{p_y}) ||_F,
\end{equation}
In our superpixel construction, the reduced image $\mathbf{\widehat{I}}$ is passed into the covariance based superpixel algorithm and a 2-D superpixel label map is generated. This map is applied back to $\mathbf{\widehat{I}}$ to obtain our 3-D superpixel mapping.

\subsection{Feature Extraction}
From each superpixel $\mathcal{S}_i$ we extract three different features. By applying a mean filter to each superpixel we can extract localised spatial information. The mean feature vector is denoted as $\vec{\mathcal{S}}^m_i$ and it is defined as 
\begin{equation}
	\vec{\mathcal{S}}^m_i = \frac{ \sum_{j=1}^{n_i} \mathbf{\widehat{I}}(p_{i,j}) }{n_i}.
\end{equation}
To obtain the spatial information surrounding a superpixel, we take a weighted combination of the information present in the adjacent superpixels. For each given superpixel $\mathcal{S}_i$, we define the set $\mathcal{Z}_i = \{ z_1 , z_2 .. , z_J \}$ which contains the $J$ indexes of the adjacent superpixels. The weighted feature vector $\vec{\mathcal{S}}_i^w$ is given by
\begin{equation}
		\vec{\mathcal{S}}_i^w = \sum_{j=1}^{J} w_{i,z_j} \vec{\mathcal{S}}_{z_j}^m,  
\end{equation}
where $h$ is a predefined scalar parameter and the weight between adjacent superpixels $w_{i,z_j}$ is defined as 
\begin{equation}
		w_{i,z_j} = \frac{ \exp \left(  -||  \vec{\mathcal{S}}_{z_j}^m - \vec{\mathcal{S}}_{i}^m ||_2^2 / h  \right) }{ \sum_{j=1}^{J} \exp \left(  -||  \vec{\mathcal{S}}_{z_j}^m - \vec{\mathcal{S}}_{i}^m ||_2^2 / h  \right)}
\end{equation}
Finally, we extract the location of the centre of each superpixel $\vec{\mathcal{S}}^p_i$ which we calculate as \begin{equation}
	\vec{\mathcal{S}}^p_i = \frac{ \sum_{j=1}^{n_i} p_{i,j} }{n_i}.
\end{equation}

\subsection{Graph Construction and SSL Classification}
Using these feature vectors we construct a weighted, undirected graphical representation $G = (V,E,W)$ where each node is a superpixel and the edge weights reflect the similarity between superpixels. Note that a similarity of 1 implies most similar and an decreasing number means less similar. The weight between superpixels is given by $w_{ij} = s_{ij}l_{ij}$,  
\begin{equation}
s_{ij} =   \exp\left( \frac{\left( \beta-1 \right)||  \vec{\mathcal{S}}^{w}_i - \vec{\mathcal{S}}^{w}_j ||_2^2  - \beta||\vec{\mathcal{S}}^m_i - \vec{\mathcal{S}}^m_j ||_2^2}{\sigma_s^2}  \right) ,
\end{equation}
\begin{equation}
l_{ij} =   \exp \left( \frac{-||\vec{\mathcal{S}}^{p}_i - \vec{\mathcal{S}}^{p}_j ||_2^2}{\sigma_l^2}\right).
\end{equation}
The parameter $\beta$ weights the contribution of the mean and weighted feature vectors and $\sigma_s , \sigma_l$ determine the width of the Gaussian kernels. The edge set is constructed using $k$-nearest neighbours.

\begin{figure*}[t]
    \centering
    \includegraphics[width=1\textwidth]{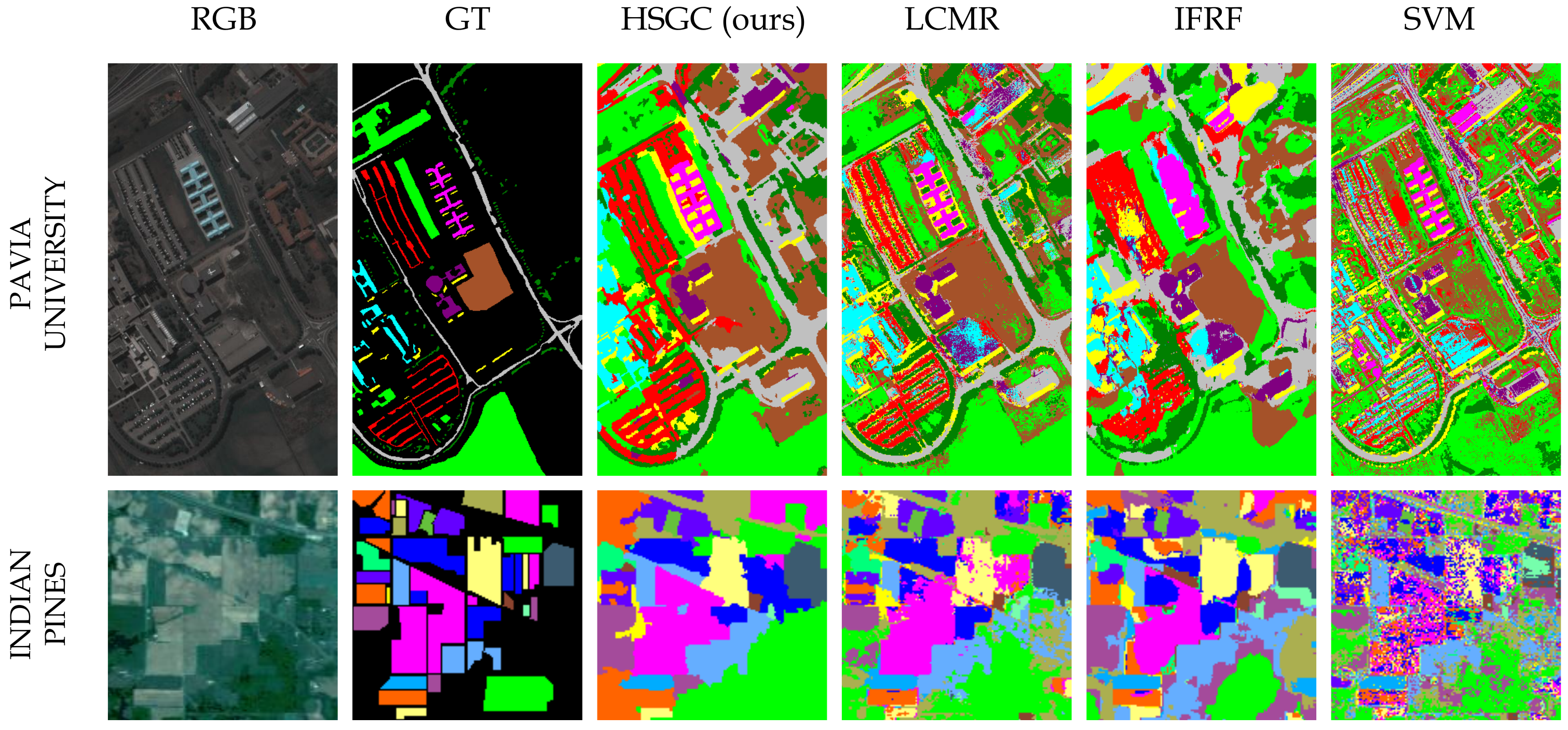} 
    \label{fig:comparisonB5}
    \caption{\textbf{The classification maps.} A comparison of the classification maps produced for the two data sets. From left to right: the RGB image, the ground truth (GT), and  classification maps from: HSGC (our approach), LCMR \cite{LCMR}, IFRF \cite{IFRF} and SVM \cite{melgani2004classification} methods when ten labelled pixels for each class are used for training. Note that HSGC and LCMR are SSL based methods. }
\end{figure*}

The initial labelling of the superpixels is specified using the matrix $Y \in \mathbb{R}^{K \times c}$, where $c$ is the number of classes present and $K$ is the number of superpixels. $Y_{vl}$ specifies the value of the seed label $l$ for node $v$. The initial label information for each superpixel is taken as the average initial label of its set of pixels. If no pixel within a superpixel is labelled the superpixel has no initial label. The graph $G$ and the labelling matrix $Y$ are then fed into the \textit{Learning with Local and Global Consistency algorithm} (LGC) by Zhou et al~\cite{LGC}. LGC propagates the labelling information across the graph and produces the final superpixel labelling matrix $F \in \mathbb{R}^{K \times c}$.  The final label for each superpixel $y_i$ is given by,
\begin{equation}
    y_i = \operatorname*{argmax}_{j \in \{1,..,c \}} F_{ij}.
\end{equation}
Each superpixel label is then passed down to its corresponding set of pixels.

\section{Experimental Results}
This section addresses the experimental methodology that we used to validate and assess our proposed approach.

\textbf{Data Description.} We use two benchmark datasets. \textit{University of Pavia:} with dimensions of $610 \times 340 \times 103$, a spectral range from $0.43$ to $0.86 \mu$m and a spatial resolution of $1.3$m. \textit{Indian Pines:} dimensions of $145 \times 145 \times 200$, a spectral range of $0.4$ to $2.5 \mu$m and a spatial resolution of $20$m. 

\textbf{Experimental Setup.} We compared our proposed approach against 
three state-of-the-art HSI classification methods: local covariance matrix representation (LCMR)~\cite{LCMR}, a SVM method~\cite{melgani2004classification}  and image fusion and recursive filtering (IFRF) \cite{IFRF}. The first method is semi-supervised and the last two are supervised methods.
The parameters of the compared approaches were set to the default values referenced in the papers. For our method the parameters were determined by an empirical coarse to fine search method. The spectral dimension after PCA was set by demanding that the total explained variance ratio was $ \geq 0.98$. The number of superpixels chosen was $1000$ for Indian Pines and $2000$ for the University of Pavia.  
To demonstrate the performance of our method in \textit{minimal supervision} settings, we only use very small training sets ranging from three to twenty labelled pixels per class. All experiments are repeated ten times and  use three common metrics to evaluate our performance: overall accuracy (OA), average accuracy (AA) and the Kappa coefficient. 
\begin{table}[t!]
    \centering
	\caption{\small{OA (\%) AA (\%) and Kappa (\%) with ten training samples per class, and for the two HSI datasets.}}
	\resizebox{\columnwidth}{!}{
	\begin{tabular}{|c|c|c|c|c|}
	\hline
\multicolumn{5}{|c|}{\cellcolor[HTML]{EFEFEF}\textsc{University of Pavia Dataset}} \\ \hline
		\hline
		 & \textbf{HSGC (ours)} & \textbf{LCMR } \cite{LCMR} & \textbf{IFRF} \cite{IFRF} & \textbf{SVM} \cite{melgani2004classification}\\ \hline
		\textbf{OA}  & \cellcolor[HTML]{9AFF99} 91.5 $\pm$ 2.6\% & 87.9 $\pm$ 3.8\% & 79.1 $\pm$ 3.8\% & 69.7 $\pm$ 3.1\% \\ \hline
		\textbf{AA}  & 90.2 $\pm$ 7.8\% & \cellcolor[HTML]{9AFF99}90.4 $\pm$ 2.6\% & 73.2 $\pm$ 3.1\% & 77.8 $\pm$ 1.3\% \\ \hline
		\textbf{Kappa}  & \cellcolor[HTML]{9AFF99}88.9 $\pm$ 3.3\% & 84.3 $\pm$ 4.7\% & 73.0 $\pm$ 4.6\% & 61.9 $\pm$ 3.3\% \\ \hline
		\multicolumn{5}{|c|}{\cellcolor[HTML]{EFEFEF}\textsc{Indian Pines Dataset}} \\ \hline \hline
		%& HSGC & LCMR \cite{LCMR} & IFRF \cite{IFRF} & SVM \cite{melgani2004classification}\\ \hline
		\textbf{OA}  & \cellcolor[HTML]{9AFF99}89.8 $\pm$ 1.7\% & 83.4 $\pm$ 2.1\% & 80.1 $\pm$ 3.2\% & 54.1 $\pm$ 2.7\% \\ \hline
		\textbf{AA}  & \cellcolor[HTML]{9AFF99}93.8 $\pm$ 7.7\% & 90.6 $\pm$ 1.5\% & 74.8 $\pm$ 2.8\% & 67.2 $\pm$ 1.6\% \\ \hline
		\textbf{Kappa}  &\cellcolor[HTML]{9AFF99} 88.4 $\pm$ 2.0\% & 81.2 $\pm$ 2.3\% & 77.5 $\pm$ 3.5\% & 48.7 $\pm$ 2.7\% \\ \hline
		%\textsc{Salinas} & HSGC & LCMR \cite{LCMR} & IFRF \cite{IFRF} & SVM \cite{melgani2004classification}\\ \hline
		%OA  & 98.9 $\pm$ 0.5\% & 94.5 $\pm$ 1.3\% & 95.6 $\pm$ 1.4\% & 82.2 $\pm$ 1.6\% \\ \hline
		%AA  & 98.6 $\pm$ 2.1\% & 96.8 $\pm$ 0.8\% & 96.4 $\pm$ 1.3\% & 89.9 $\pm$ 1.1\% \\ \hline
		%Kappa  & 98.8 $\pm$ 0.5\% & 93.8 $\pm$ 1.5\% & 95.1 $\pm$ 1.6\% & 80.2 $\pm$ 1.8\% \\ \hline
	\end{tabular}
	}
	\label{table1}
\end{table}

\textbf{Comparison With Other Methods.}
we start by visually evaluating our approach against the three different classifiers. The results are visualised in Fig.1. We can observe that our method produces a smoother output, with significantly fewer outliers than the compared approaches. It deals well with the complex structures present in the Pavia data set and preserves the boundaries between the different regions in Indian Pines.

\begin{figure}
    \centering
    \includegraphics[width=0.48\textwidth]{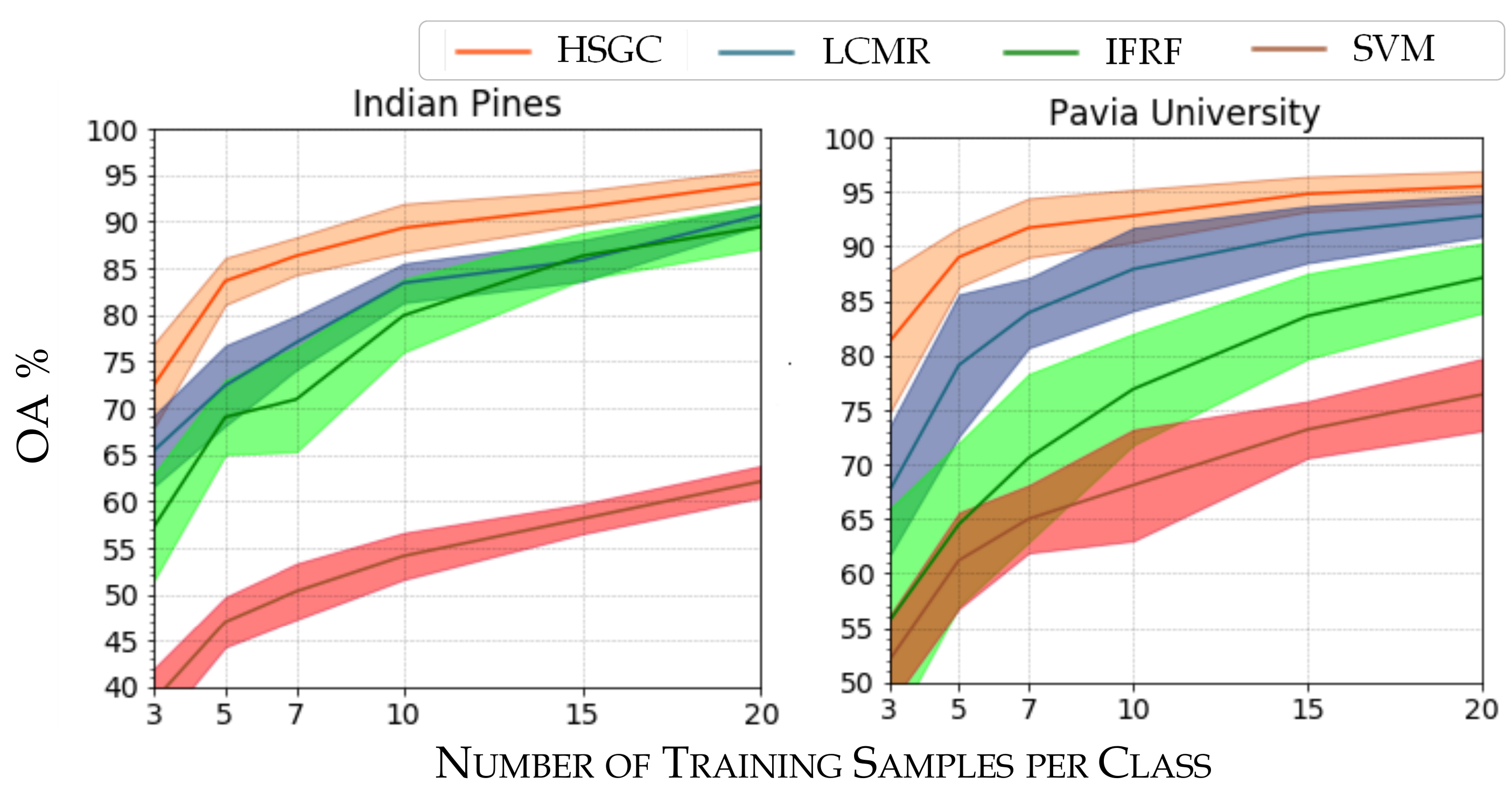}
    \caption{\textbf{Overall Classification Accuracy (OA).} The lines represent the average data for ten runs whilst the shaded regions represent the standard deviation of: LCMR \cite{LCMR}, IFRF \cite{IFRF}, SVM \cite{melgani2004classification} and the proposed HSGC.}
        \label{fig:updatedPlot}
\end{figure}

This is reflected in the numerical results reported in Table~\ref{table1}, where the OA, the AA and Kappa coefficient were calculated  using \textit{ten} labelled pixels per class. We observe that our approach outperformed the other methods for both datasets. However, we can observe in Fig.~\ref{fig:updatedPlot} and Table~\ref{pavia} that our approach \textit{significantly} outperforms the compared methods when the number of labelled pixels per class is reduced below 10. We can observe, that for all number of labels counts and for both datasets, our approach exhibits the highest overall classification accuracy (OA). Note that higher values of OA correspond to better classifier performance. Overall, our approach outperformed the compared state-of-the-art approaches. The main contribution of our approach is to produce very high classification accuracy even when the number of labelled pixels per class is incredibly small.
\begin{table}[t!]
    \centering
	\caption{\small{OA (\%) on the two benchmarking datasets.}}
	\resizebox{\columnwidth}{!}{
	\begin{tabular}{|c|c|c|c|c|}
	\hline
	\multicolumn{5}{|c|}{\cellcolor[HTML]{EFEFEF}\textsc{University of Pavia Dataset}} \\ \hline
		\hline
		%\multicolumn{1}{|c|}{University of Pavia} &
		%\multicolumn{4}{l|}{Semi-Supervised Technique} \\
		%\hline		
		Labels$\setminus$ Class & \textbf{HSGC (ours)} & \textbf{LCMR} \cite{LCMR} & \textbf{IFRF} \cite{IFRF} & \textbf{SVM} \cite{melgani2004classification}\\ \hline
		3  & \cellcolor[HTML]{9AFF99}81.2 $\pm$ 6.5\% & 67.5 $\pm$ 5.9\% & 55.6 $\pm$ 10.3\% & 52.0 $\pm$ 4.0\% \\ \hline
		7  & \cellcolor[HTML]{9AFF99}91.7 $\pm$ 2.7\% & 83.9 $\pm$ 3.2\% & 70.6 $\pm$ 7.7\% & 65.0 $\pm$ 3.1\% \\ \hline
		20 & \cellcolor[HTML]{9AFF99}95.5 $\pm$ 1.4\% & 92.8 $\pm$ 1.9\% & 87.1 $\pm$ 3.2\% & 76.4 $\pm$ 3.3\% \\ \hline
		\multicolumn{5}{|c|}{\cellcolor[HTML]{EFEFEF}\textsc{Indian Pines Dataset}} \\ \hline \hline
		3  & \cellcolor[HTML]{9AFF99}72.4 $\pm$ 4.5\% & 65.4 $\pm$ 3.8\% & 57.2 $\pm$ 5.7\% & 38.5 $\pm$ 3.5\% \\ \hline
		7  & \cellcolor[HTML]{9AFF99}84.3 $\pm$ 2.0\% & 77.0 $\pm$ 2.9\% & 70.9 $\pm$ 5.6\% & 50.3 $\pm$ 3.0\% \\ \hline
		20 & \cellcolor[HTML]{9AFF99}94.1 $\pm$ 1.5\% & 90.7 $\pm$ 1.1\% & 89.4 $\pm$ 2.3\% & 62.1 $\pm$ 1.7\% \\ \hline
	\end{tabular}
	}
	\label{pavia}
\end{table}
\vspace{-2.5mm}
\section{Conclusion}
In this work, we present a novel framework, HSGC, for hyperspectral image classification. Our framework combines a novel, purpose built superpixel algorithm with a semi-supervised graph based approach. We demonstrate state-of-the-art results compared with a recent semi-supervised and two supervised approaches. Our highlight is that HSGC produces the highest classification accuracy, even when the amount of labelled data is small. This shows the benefits and potential of learning with minimal supervision on graphs.
\bibliographystyle{IEEEbib}
\bibliography{bibliographyV2.bib}

\end{document}